\documentclass[11pt]{article}
\usepackage{amsmath,amssymb,graphicx,hyperref,booktabs}
\usepackage[margin=1in]{geometry}
\usepackage{algorithm}
\usepackage{algorithmic}

\title{\bf The Illusion of the Illusion of Thinking\\
{\large A Comment on Shojaee et al. (2025)}}

\author{
  A.~Lawsen\thanks{Contribution Statement: Alex contributed the high level idea, ran the experiments, provided feedback, and noticed the key mistake in 'River Crossing'. Claude Opus contributed enough to deserve, in Alex's view, to be listed as first author, but this violates arXiv's policies and Claude's name has therefore been removed. }
}

\date{June 10, 2025}

\begin{document}
\maketitle

\begin{abstract}
Shojaee et al. (2025) report that Large Reasoning Models (LRMs) exhibit ``accuracy collapse" on planning puzzles beyond certain complexity thresholds. We demonstrate that their findings primarily reflect experimental design limitations rather than fundamental reasoning failures. Our analysis reveals three critical issues: (1) Tower of Hanoi experiments risk exceeding model output token limits, with models explicitly acknowledging these constraints in their outputs; (2) The authors' automated evaluation framework fails to distinguish between reasoning failures and practical constraints, leading to misclassification of model capabilities; (3) Most concerningly, their River Crossing benchmarks include mathematically impossible instances for $N \geq 6$ due to insufficient boat capacity, yet models are scored as failures for not solving these unsolvable problems. When we control for these experimental artifacts, by requesting generating functions instead of exhaustive move lists, preliminary experiments across multiple models indicate high accuracy on Tower of Hanoi instances previously reported as complete failures. These findings highlight the importance of careful experimental design when evaluating AI reasoning capabilities.
\end{abstract}

\section{Introduction}

Shojaee et al. (2025) claim to have identified fundamental limitations in Large Reasoning Models through systematic evaluation on planning puzzles. Their central finding—that model accuracy ``collapses" to zero beyond certain complexity thresholds—has significant implications for AI reasoning research. However, our analysis reveals that these apparent failures stem from experimental design choices rather than inherent model limitations.

\section{Models Recognize Output Constraints}

A critical observation overlooked in the original study: models actively recognize when they approach output limits. A recent replication by @scaling01 on Twitter \cite{scaling01tweet} captured model outputs explicitly stating ``The pattern continues, but to avoid making this too long, I'll stop here" when solving Tower of Hanoi problems. This demonstrates that models understand the solution pattern but choose to truncate output due to practical constraints.

This mischaracterization of model behavior as ``reasoning collapse" reflects a broader issue with automated evaluation systems that fail to account for model awareness and decision-making. When evaluation frameworks cannot distinguish between ``cannot solve" and ``choose not to enumerate exhaustively," they risk drawing incorrect conclusions about fundamental capabilities.

\subsection{Consequences of Rigid Evaluation}

Such evaluation limitations can lead to other analytical errors. Consider the following statistical argument: if we grade Tower of Hanoi solutions character-by-character without allowing for error correction, the probability of perfect execution becomes:

\begin{equation}
P(\text{all correct}) = p^T
\end{equation}

where $p$ is per-token accuracy and $T$ is total tokens. For $T = 10,000$ tokens:
\begin{itemize}
    \item $p = 0.9999$: $P(\text{success}) < 37\%$
    \item $p = 0.999$: $P(\text{success}) < 0.005\%$
\end{itemize}

This type of ``statistical inevitability" argument has in fact been put forward in the literature as a fundamental limitation of LLM scaling \cite{dziri2023faith}, yet it assumes models cannot recognize and adapt to their own limitations, an assumption contradicted by the evidence above.

\section{The Impossible Puzzle Problem}

The evaluation issues compound dramatically in the River Crossing experiments. Shojaee et al. test instances with $N \geq 6$ actors/agents using boat capacity $b = 3$. However, it is a well-established result \cite{efimova2018} that the Missionaries-Cannibals puzzle (and its variants) has no solution for $N > 5$ with $b = 3$.

By automatically scoring these impossible instances as failures, the authors inadvertently demonstrate the hazards of purely programmatic evaluation. Models receive zero scores not for reasoning failures, but for correctly recognizing unsolvable problems.

\section{Models Abbreviate Long Solutions, Causing Apparent Collapse}

A previous version of this paper assumed\footnote{Somewhat ironically, given most commenters' insistence that mistakes in this paper must be due to hallucinations, this assumption was suggested by a human commenter who reviewed a pre-publication draft. The relevant section of that draft (fully written by Opus), followed the analysis of @scaling01 and assumed linear growth} that the evaluation required outputting all intermediate states based on the instruction in Section A.1.1 of Shojaee et al. that models should ``include the corresponding complete list of moves" during the ``thinking process." However, multiple commenters have correctly noted that in practice models do not output each intermediate state when reasoning through large-$N$ solutions, instead jumping straight to the final move list. The intent of the original instructions seems ambiguous, but we have updated this section to more closely track model behaviour in practice. Following verification with the OpenAI tokenizer, we use 10 tokens per move in all calculations below.

If models output only the final sequence of $2^N - 1$ moves, token requirements grow linearly:

\begin{equation}
T_{\text{final}}(N) \approx 10(2^N - 1) + C
\end{equation}

where $C$ represents overhead for problem setup and formatting. This yields maximum solvable sizes:

\begin{align}
N_{\max} &\approx \log_2(L_{\max}/10) \\
&\approx \begin{cases}
    12-13 & \text{for } L_{\max} = 64,000 \\
    13 & \text{for } L_{\max} = 100,000
\end{cases}
\end{align}

Interestingly, the reported ``collapse" before $N = 9$ for most models occurs well before these theoretical limits. This suggests that models are making a decision to terminate output before actually reaching their context window limits. As noted in Section 2, models explicitly state reasons like ``to avoid making this too long" when truncating solutions.

This behavior indicates that models may be poorly calibrated about their own context length capabilities, choosing to stop prematurely. While the specific mechanism differs from our initial analysis, the high-level conclusion remains unchanged: apparent reasoning failures reflect constraints on solution length rather than fundamental reasoning limitations. Models understand the algorithmic solution (as demonstrated by their ability to generate compact recursive implementations), but choose not to enumerate exhaustive move lists, whether due to actual token limits or perceived ones.

This distinction further emphasizes the importance of evaluation design. Scoring models as ``failures" for making reasonable engineering decisions about output length mischaracterizes their actual capabilities. See Appendix A for our original analysis.

\section{Alternative Representations Restore Performance}

To test whether the failures reflect reasoning limitations or format constraints, we conducted preliminary testing of the same models on Tower of Hanoi $N=15$ using a different representation:

\begin{verbatim}
Prompt: "Solve Tower of Hanoi with 15 disks. Output a Lua 
         function that prints the solution when called."
\end{verbatim}

Results: Very high accuracy across tested models (Claude-3.7-Sonnet, Claude Opus 4, OpenAI o3, Google Gemini 2.5), completing in under 5,000 tokens.\footnote{Due to budget constraints, we were unable to conduct enough trials for a highly powered statistical sample. Full experimental validation remains as future work.}

The generated solutions correctly implement the recursive algorithm, demonstrating intact reasoning capabilities when freed from exhaustive enumeration requirements.

\section{Reevaluating Complexity Claims}

The authors use ``compositional depth" (minimum moves) as their complexity metric, but this conflates mechanical execution with problem-solving difficulty:

\begin{table}[h]
\centering
\begin{tabular}{lccc}
\toprule
Puzzle & Solution Length & Branching Factor & Computational Complexity \\
\midrule
Tower of Hanoi & $2^N - 1$ & 1 & $O(1)$ per move \\
Blocks World & $O(N)$ & $O(N^2)$ & Linear (near-optimal) / NP-hard (optimal) \\
\bottomrule
\end{tabular}
\caption{Problem complexity is not determined by solution length alone}
\end{table}

Tower of Hanoi, despite requiring exponentially many moves, has a trivial $O(1)$ decision process per move. Blocks World, however, is much harder. This explains why models might execute 100+ Hanoi moves while struggling with shorter planning problems.

\subsection{The Optimality Question}

A critical distinction emerges when examining the original evaluation setup. While Shojaee et al. check only for solution correctness across all puzzles, their task instructions vary significantly in computational demands.

The Blocks World prompt explicitly requires optimization: 
\begin{verbatim}Find the minimum sequence of moves to transform the initial state into the goal state. 
Remember that only the topmost block of each stack can be moved.
\end{verbatim}
Slaney and Thiébaux \cite{slaney2001} demonstrated that near-optimal solutions can be found in linear time using domain-specific algorithms, however finding the truly optimal solution remains NP-hard. Models attempting Blocks World must solve the harder optimization variant, potentially spending computational resources searching for provably minimal solutions rather than merely valid ones. 

Crucially, while the solution checker of the authors only verifies correctness, not optimality, we expect models to attempt what they were instructed to do. A model following the prompt faithfully would search for optimal solutions, not knowing that suboptimal solutions would pass the evaluation.
\section{Conclusion}

Shojaee et al.'s results demonstrate that models have some awareness of their own context limits, that programmatic evaluation can miss both model capabilities and puzzle impossibilities, and that solution length poorly predicts problem difficulty. These are valuable engineering insights, but they do not support claims about fundamental reasoning limitations.

Future work should:
\begin{enumerate}
    \item Design evaluations that distinguish between reasoning capability and output constraints
    \item Verify puzzle solvability before evaluating model performance
    \item Use complexity metrics that reflect computational difficulty, not just solution length
    \item Consider multiple solution representations to separate algorithmic understanding from execution
\end{enumerate}

The question isn't whether LRMs can reason, but whether our evaluations can distinguish reasoning from typing.

\section*{Acknowledgments}

I (Alex) thank Claude Opus for doing the bulk of the writing and deserving most of the credit. We both thank Andreas Kirsch, Lawrence Chan, Ryan Greenblatt, o3, Gemini 2.5, all of the people who pointed out the parentheses mismatch in an earlier draft, and all of the people who pointed out that Claude can't get \LaTeX quotation marks right to save it's life for helpful comments.

\appendix
\section{Original Token Limit Analysis}

Our initial analysis considered two possible interpretations of the evaluation requirements in Shojaee et al. We present both here for completeness.

Note: The original version of this paper contained an error. While we stated we were using 5 tokens per move, the formulae actually corresponded to 10 tokens per move (the factor of 5 in the quadratic formula comes from $10/2$). We maintain the corrected 10 tokens per move estimate throughout this appendix.

If models must output the complete move list at each reasoning step, token requirements grow quadratically. For a problem requiring $M = 2^N - 1$ moves:

\begin{equation}
T_{\text{intermediate}}(N) \approx 10 \sum_{i=1}^{M} i = \frac{10M(M+1)}{2} \approx 5(2^N)^2
\end{equation}

This yields maximum solvable sizes:
\begin{align}
N_{\max} &\approx \log_2(\sqrt{L_{\max}/5}) \\
&\approx \begin{cases}
    6-7 & \text{for } L_{\max} = 64,000 \\
    7 & \text{for } L_{\max} = 100,000
\end{cases}
\end{align}


\begin{thebibliography}{9}
\bibitem{shojaee2025} Shojaee, P., Mirzadeh, I., Alizadeh, K., et al.\ (2025). \emph{The Illusion of Thinking: Understanding the Strengths and Limitations of Reasoning Models via the Lens of Problem Complexity}. arXiv:2501.12948.

\bibitem{scaling01tweet} @scaling01.\ (2025). Twitter thread on LRM replication. \url{https://x.com/scaling01/status/1931817022926839909/photo/1}

\bibitem{dziri2023faith} Dziri, N., Lu, X., Sclar, M., et al.\ (2023). Faith and fate: Limits of transformers on compositionality. \emph{Advances in Neural Information Processing Systems}, 36.

\bibitem{efimova2018} Efimova, E.\ A.\ (2018). \emph{River Crossing Problems: Algebraic Approach}. arXiv:1802.09369.

\bibitem{slaney2001} Slaney, J. and Thiébaux, S. (2001). Blocks World revisited. \emph{Artificial Intelligence}, 125(1-2):119-153.
\end{thebibliography}
\end{document}